\documentclass[runningheads]{llncs}
\usepackage{graphicx}
\usepackage{comment}
\usepackage{amsmath,amssymb} 
\usepackage{color}

\usepackage{algorithm}
\usepackage[noend]{algpseudocode}

\usepackage{tabularx} 
\usepackage{booktabs}

\usepackage[inline]{enumitem} 
\usepackage{hyperref}

\usepackage{macros}

\begin{document}
\pagestyle{headings}
\mainmatter
\def\ECCVSubNumber{3241}  

\title{Solving the Blind Perspective-n-Point Problem End-To-End With Robust Differentiable Geometric Optimization}

\titlerunning{Solving the Blind Perspective-n-Point Problem End-To-End}
\author{Dylan Campbell$^\star$
	\and
	Liu Liu\thanks{Equal contribution; corresponding author email: \texttt{dylan.campbell@anu.edu.au}}
	\and
	Stephen Gould
}
\authorrunning{D. Campbell et al.}
\institute{Australian National University, Australian Centre for Robotic Vision\thanks{This research was conducted by the Australian Research Council Centre of Excellence for Robotic Vision (CE140100016) and funded by the Australian Government.}
}

\maketitle

\begin{abstract}
Blind Perspective-n-Point (PnP) is the problem of estimating the position and orientation of a camera relative to a scene, given 2D image points and 3D scene points, without prior knowledge of the 2D--3D correspondences. Solving for pose and correspondences simultaneously is extremely challenging since the search space is very large. Fortunately it is a coupled problem: the pose can be found easily given the correspondences and vice versa. Existing approaches assume that noisy correspondences are provided, that a good pose prior is available, or that the problem size is small. We instead propose the first fully end-to-end trainable network for solving the blind PnP problem efficiently and globally, that is, without the need for pose priors. We make use of recent results in differentiating optimization problems to incorporate geometric model fitting into an end-to-end learning framework, including Sinkhorn, RANSAC and PnP algorithms. Our proposed approach significantly outperforms other methods on synthetic and real data.
\end{abstract}

\section{Introduction}
\label{sec:introduction}

\begin{figure*}[!t]\centering
	\includegraphics[width=0.98\textwidth]{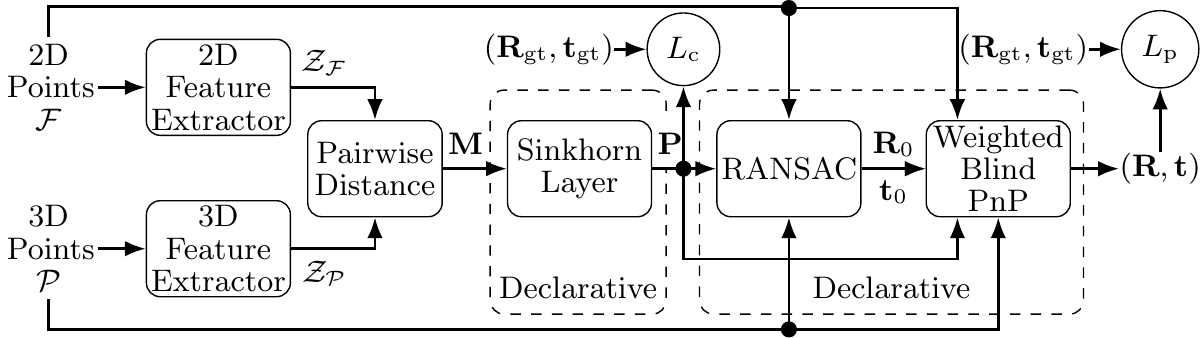}
	\caption{
		$\!$Network architecture for our end-to-end blind PnP solver.
		We combine standard neural layers with declarative layers in a bi-level optimization framework to instantiate the traditional camera pose estimation pipeline (feature extraction, feature matching and optimization) in a single neural network.
		The input is a set of 2D and 3D point coordinates, from which point-wise features are extracted using standard network layers.
		Feature matching is then performed by computing the pairwise distance between the 2D and 3D point features, and using the Sinkhorn algorithm \cite{sinkhorn1967diagonal} to obtain a joint probability matrix.
		Finally, a probability-weighted blind PnP objective function is optimized from a RANSAC initialization to estimate the camera rotation and translation.
		The key contribution of this work is showing how this optimization procedure may be incorporated into an end-to-end learnable network by the use of declarative layers.
	}
	\label{fig:flowchart}
\end{figure*}

The blind Perspective-n-Point (PnP) problem \cite{moreno2008pose} aims to estimate the camera pose from which a set of 2D points were viewed, relative to an unordered 3D point-set.
Specifically, the task is to find the rotation and translation that aligns a set of 2D bearing vectors with a set of 3D points, without knowledge of the true 2D--3D correspondences.
The camera intrinsic parameters are assumed to be known, which allows 2D points to be expressed as bearing vectors.
While a fundamental technique for many computer vision and robotic applications, including augmented reality and visual localization, it remains a challenging problem that has not as yet been satisfactorily solved.

The standard (non-blind) PnP problem \cite{fischler1981random,lepetit2009epnp}, where 2D--3D correspondences are known, is significantly less complex.
It has a closed-form solution for three points \cite{kneip2011novel} and, for a larger number of points, can be embedded in a RANSAC framework \cite{fischler1981random} to reduce its sensitivity to outliers.
However, it is inherently difficult to establish 2D--3D correspondences between modalities.
As a result, PnP solvers are typically restricted to applications where both the 2D and 3D data contain visual information, such as structure-from-motion datasets \cite{li2018megadepth}.
Even for these, appearance may change seasonally, diurnally, and with weather, and so using geometric rather than visual features may improve generalizability.

Solving the PnP problem without correspondences is much more challenging, because the search space of correspondences and camera poses is very large, the objective function has many local optima, and outliers are prevalent.
As a result, it was traditionally the domain of robust geometric algorithms that overcame the search space and non-convexity problems by requiring good pose priors \cite{david2004softposit,moreno2008pose} or time-consuming global optimization \cite{brown2015globally,campbell2018globally,campbell2019alignment}.
Since these techniques were typically iterative, randomized and non-differentiable, this problem has not been amenable to a deep learning solution.
Moreover, the geometry of the problem is difficult for a network to learn.
However, there is significant opportunity in using a neural network for this problem, since it can effectively recognize patterns in the geometric data and thus reduce the search space and influence of outliers.

Fortunately, the framework of deep declarative networks \cite{gould2019deep} has recently been proposed, which provides a solution to the problem of including standard neural layers and geometric optimization layers inside the same end-to-end learnable network.
This paper applies many of the ideas associated with deep declarative networks by formulating our deep blind PnP solver as a bi-level optimization problem.
In this way, we aim to benefit from the pattern recognition capabilities of standard neural networks and the physical models and optimization algorithms used in traditional geometric approaches to the PnP problem.

We focus on the \emph{optimization} part of the traditional camera pose estimation pipeline (feature extraction, matching and optimization) shown in \figref{fig:flowchart}, that is, the remit of the PnP solver itself.
To this end, we use an existing network architecture for feature extraction \cite{yi2018learning} and matching \cite{liu2019deep}.
However, our key insight is that camera pose optimization algorithms, including robust global search techniques such as RANSAC \cite{fischler1981random} and state-of-the-art nonlinear PnP solvers, can be seamlessly integrated into an end-to-end deep learning framework.
Our contributions are:
\begin{enumerate*}[label=\arabic*)]
	\item the first fully end-to-end trainable network for solving the blind PnP problem efficiently and globally;
	\item the novel deployment of geometric model fitting algorithms as declarative layers inside the network;
	\item the novel embedding of non-differentiable robust estimation techniques into the network; and
	\item state-of-the-art performance on synthetic and real datasets.
\end{enumerate*}

\section{Related Work}
\label{sec:related_work}

The majority of camera pose estimation methods assume that a set of 2D--3D correspondences is available and thus a PnP solver \cite{fischler1981random,lepetit2009epnp} can be used.
Hence, much of the effort in improving the visual localization pipeline focuses on robustly establishing correspondences \cite{sattler2017efficient,schonberger2018semantic} or removing outliers \cite{enqvist2008robust,svarm2016city,yi2018learning}.
These approaches are not appropriate for situations where correspondences cannot be easily obtained, such as when the 3D point-set has no associated visual information.
In contrast, we address this problem by deferring the correspondence estimation task until the PnP stage of the pipeline, jointly estimating pose and correspondences.
An alternative approach, which does not require explicit correspondences, is learning-based pose regression \cite{kendall2015posenet,kendall2017geometric,walch2017image}.
However, Sattler \etal \cite{sattler2019understanding} show that this essentially solves an image retrieval task rather than reasoning about 3D structure.
Also, the camera is not localized with respect to an explicit 3D map, instead representing the scene implicitly.
DSAC and extensions \cite{brachmann2017dsac,brachmann2018learning} can localize with respect to a 3D model, but require many training images from the test scene.
In contrast, we eschew visual information to learn generalizable geometric features, and never see the test scene during training.

To solve the blind PnP problem \cite{moreno2008pose} of localizing the camera relative to an explicit unseen 3D model when correspondences cannot be obtained,
some approaches seek a local optimum and assume that a good pose prior is available \cite{david2004softposit,baka2014oriented}.
For example, David \etal \cite{david2004softposit} propose an algorithm that alternates between solving for pose and correspondences, using the Sinkhorn algorithm \cite{sinkhorn1967diagonal}.
To mitigate the pose prior requirement, global search strategies have been proposed, including multiple random starts \cite{david2004softposit} and probabilistic pose priors \cite{moreno2008pose}.
RANSAC \cite{fischler1981random} can also be used, but becomes intractable for moderately-sized problems.
To obviate the need for pose priors and guarantee that a global optimum is found, globally-optimal approaches \cite{campbell2018globally,brown2015globally,campbell2019alignment} use branch-and-bound to systematically reduce the search space.
For example, Campbell \etal \cite{campbell2019alignment} globally optimize a robust distance between mixture distributions to solve the blind PnP problem.
However, these optimal methods are time-consuming and limited to a moderate number of points, unlike our (orders of magnitude) faster approach.

Deep PnP solvers are proposed in existing work \cite{dang2018eigendecomposition} (standard PnP) and concurrent work \cite{liu2019deep} (blind PnP).
Due to difficulties inherent in eigendecomposition and outlier filtering, neither approach is end-to-end, despite being highly effective at learning 2D--3D correspondences.
The former shows how to avoid unstable eigendecomposition gradients by applying a loss before the pose parameters are estimated,
while the latter shows that high-quality 2D--3D correspondence matrices can be learned using optimal transport via the Sinkhorn algorithm \cite{sinkhorn1967diagonal}.
Metric learning can also be used to learn matchable features, as shown for 2D--2D and 3D--3D matching \cite{fathy2018hierarchical}.
However, estimating pose from these features or correspondence matrices requires a non-differentiable selection step, such as nearest neighbor search, to reduce the set of correspondences to a tractable size.
Different to these approaches, we propose a fully end-to-end trainable blind PnP solver.
We directly use an existing ResNet-based \cite{he2016deep} feature extraction architecture \cite{yi2018learning} and a Sinkhorn-based \cite{sinkhorn1967diagonal} feature matching technique \cite{liu2019deep} in our network, since our focus is on the joint optimization of all parameters in the camera pose estimation pipeline.
Our contribution is orthogonal to these works.

The declarative framework that allows us to incorporate geometric optimization algorithms into a deep network is described in Gould \etal \cite{gould2019deep}.
They present theoretical results and analyses on how to differentiate constrained optimization problems via implicit differentiation.
Differentiable convex problems have also been studied recently, including quadratic programs \cite{amos2017optnet} and cone programs \cite{agrawal2019differentiating,agrawal2019differentiable}.
In computer vision, the technique has been applied to video classification \cite{fernando2016learning,fernando2017discriminatively}, action recognition \cite{cherian2017generalized}, visual attribute ranking \cite{santacruz2018visual}, few-shot learning for visual recognition \cite{lee2019meta}, and non-blind PnP in concurrent work \cite{chen2020end}.
In this work, we show that we can embed geometric model fitting algorithms and non-differentiable robust estimation techniques into a network as declarative layers to solve the blind PnP problem end-to-end.

\section{An End-To-End Blind PnP Solver}
\label{sec:method}

In this section we present our end-to-end trainable network for solving the blind PnP problem, which we name BPnPNet.
We start by formally defining the problem, then provide background on deep declarative networks and show how critical components of a blind PnP solver can be implemented as declarative layers.
We then describe our network architecture, loss functions, and learning strategy.

\subsection{Problem Formulation}
\label{sec:method_formulation}

Let $\bp \in \reals^3$ denote a 3D point and $\bbf \in \reals^3$ denote a unit bearing vector corresponding to a 2D point in the image plane of a calibrated camera.
That is, $\|\bbf\| = 1$ and $\bbf \propto \bK^{-1}[u, v, 1]\transpose$, where $\bK$ is the matrix of intrinsic camera parameters and $(u, v)$ are the 2D image coordinates.
Given a set of bearing vectors $\cF = \{\bbf_i\}_{i=1}^{m}$ and 3D points $\cP = \{\bp_i\}_{i=1}^{n}$, the objective of blind PnP is to find the rotation $\bR \in SO(3)$ and translation $\bt \in \reals^3$ that transforms $\cP$ to the coordinate system of $\cF$ with the greatest number of one-to-one inlier correspondences, defined by an angular inlier threshold $\theta \in (0, \pi)$.
The optimization problem is
\pagebreak[0]
\begin{equation}
\label{eqn:optimisation_problem}
\arraycolsep=2pt
\begin{array}{>{\displaystyle}r>{\displaystyle}l}
\maximize_{\bR, \bt, \bC} & 
\sum_{i=1}^{m}\sum_{j=1}^{n}  \bC_{ij}  \left( 2 \bigind{\angle \left(\bbf_i, \bR\bp_j + \bt \right) \leqslant \theta} - 1\right)\\
\st & \bR \in SO(3), \; \bt \in \reals^3\\
& \bC \in \bools^{m \times n}, \; \bC\ones^n \in \bools^m, \; \bC\transpose\ones^m \in \bools^n
\end{array}
\end{equation}
where $\bC$ is a Boolean one-to-one correspondence matrix with at most one non-zero element in each row and column, $\bC_{ij}$ is the element at row $i$ and column $j$, $\bbB = \{0, 1\}$ is the Boolean domain, \ind{\!\cdot\!} is an Iverson bracket, and $\angle(\bx, \by) = \arccos(\|\bx\|^{-1}\|\by\|^{-1} \bx\transpose \by)$ is a function that returns the angle in $[0, \pi]$ between the vector arguments.
This inlier maximization formulation optimizes a robust angular reprojection error.
The joint optimization problem can be simplified if either the correspondences or the camera pose is known.
If the correspondence matrix $\bC$ is known, we have the standard PnP problem. The camera pose can be estimated by minimizing the angular reprojection error
\begin{equation}
\label{eqn:reprojection_error}
\frac{1}{mn} \sum_{i=1}^{m}\sum_{j=1}^{n} \bC_{ij} \angle \left(\bbf_i, \bR\bp_j + \bt \right).
\end{equation}
If rotation $\bR$ and translation $\bt$ are known, then $\bC$ can be computed as
\begin{equation}
\label{eqn:correspondences_given_pose}
\widetilde{\bC}_{ij} = \bigind{\angle \left(\bbf_i, \bR\bp_j + \bt \right) \leqslant \theta}
\end{equation}
followed by the Hungarian algorithm to enforce the one-to-one constraint.
If a good pose or correspondence matrix initialization is available, these steps can be alternated to find a good estimate of $\bR$, $\bt$ and $\bC$.
This strategy, analogous to the Iterative Closest Point algorithm~\cite{besl1992method}, is taken by the SoftPOSIT algorithm~\cite{david2004softposit}.
In contrast, our approach applies this alternation implicitly during training.

\subsection{Bi-Level Optimization}
\label{sec:method_bilevel}

\begin{figure}[!t]\centering
	\includegraphics[width=\textwidth]{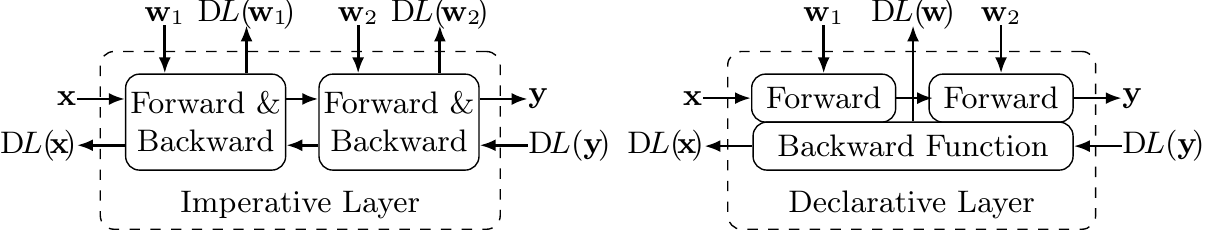}
	\caption{
		Comparison of imperative and declarative layers.
		An imperative layer (left) transforms the input $\bx$ to the output $\by$ using explicit forward functions, parameterized by network weights $\bw$.
		A declarative layer (right) computes the output $\by$ as a minimizer of an objective function, parameterized by the input $\bx$ and network weights $\bw$.
		During learning, the gradient $\dd L(\by)$ of the global loss function with respect
		to the output is propagated backwards using the chain rule.
		While the backward function of an imperative layer is tightly coupled with every step of the forward function, effectively unrolling any algorithm applied, the backward function of a declarative layer computes the gradient of the entire layer in one step. The individual forward processing nodes can be recursive or non-differentiable, provided that the objective function optimized by the final forward node is (sub)differentiable in $\bx$.}
	\label{fig:layer_comparison}
\end{figure}

The conceptual framework that underpins this method is the deep declarative network \cite{gould2019deep}, which interprets training a network as a bi-level optimization problem.
According to this view, a network can be composed of multiple imperative and declarative layers.
An imperative layer explicitly defines a function for transforming the input to the output, \eg, a convolution layer.
In contrast, a declarative layer is implicitly defined in terms of the desired output, formulated as a constrained mathematical optimization problem.
Declarative layers are more flexible and general than standard layers, since they admit constraints on the output and decouple the gradient computation from the algorithm used to solve the optimization problem.
Crucially, the technique of implicit differentiation enables the back-propagation of gradients through a declarative layer without having to traverse the forward processing function, as shown in \figref{fig:layer_comparison}.

A bi-level optimization problem \cite{stackelberg2011market,bard1998practical} for end-to-end learning has an upper-level problem solved subject to constraints imposed by a lower-level problem:
\begin{align}
\begin{array}{ll}
\minimize & L(\bx, \by) \\
\st & \by \in \argmin_{\bu \in \cC} f(\bx, \bu)
\end{array}
\label{eqn:bilevel}
\end{align}
where $L$ is a global loss function, $f$ is an objective function, $\cC$ is an arbitrary constraint set, and the loss is minimized over all network weights.
To solve the bi-level optimization problem \eqnref{eqn:bilevel} by gradient descent, we require the derivative
\begin{align}
\dd L(\bx, \by) &= \dd[X] L(\bx, \by) + \dd[Y] L(\bx, \by) \dd \by(\bx)
\end{align}
in order to back-propagate gradients.
The key challenge is to compute $\dd \by(\bx)$, for which we use implicit differentiation.
We use the notation $\dd f$ for the derivative of a function $f : \reals^n \to \reals^m$, an $m \times n$ matrix with entries $(\dd f(\bx))_{ij} = \partial f_i / \partial \bx_j$,
and we denote the partial derivative over the formal variable $X$, with all other variables fixed, as $\dd[X] f(\bx, \by)$.
We also use $\dd[XY]^2 f$ as shorthand for $\dd[X] (\dd[Y] f)\transpose$.
For completeness, we collect the two results from Gould \etal \cite{gould2019deep} that are used in this paper.
The unconstrained case applies Dini's implicit function theorem~\cite[p19]{dontchev2014implicit} to the first-order optimality condition $\dd[Y] f(\bx, \by) = \zeros$.
\begin{lemma}
	\label{lm:unconstrained}
	Consider a function $f: \reals^n \times \reals^m \to \reals$ and let $\by(\bx) \in \argmin_{\bu} f(\bx, \bu)$.
	Assume $\by(\bx)$ exists and that $f$ is second-order differentiable
	in the neighborhood of $\bu = \by(\bx)$.
	Set $\bH = \dd[YY]^2 f(\bx, \by(\bx)) \in \reals^{m \times m}$ and $\bB = \dd[XY]^2 f(\bx, \by(\bx)) \in \reals^{m \times n}$.
	Then for non-singular $\bH$ the derivative of $\by$ with respect to $\bx$ is
	\begin{equation}
	\label{eqn:unconstrained}
	\dd \by(\bx) = -\bH^{-1} \bB.
	\end{equation}
\end{lemma}
We also require the linear equality constraints case.
\begin{lemma}
	\label{lm:linear_equality}
	Consider a function $f: \reals^n \times \reals^m \to \reals$ and
	let $\bA \in \reals^{p \times m}$ and $\bd \in \reals^p$ with $\text{rank}(\bA) = p$
	define a set of $p$ under-constrained linear equations $\bA\bu = \bd$.
	Also let $\by(\bx) \in \argmin_{\bu} f(\bx, \bu)$ subject to $\bA\bu = \bd$.
	Assume that $\by(\bx)$ exists and that $f(\bx, \bu)$ is second-order differentiable in the neighborhood of $\bu = \by(\bx)$.
	Set $\bH = \dd[YY]^2 f(\bx, \by)$ and $\bB = \dd[XY]^2 f(\bx, \by)$.
	Then
	\begin{equation}
	\label{eqn:linear_equality}
	\dd \by(\bx) = \left(\bH^{-1}\!\bA\!\transpose (\bA\bH^{-1} \bA\!\transpose)^{-1}\!\bA\bH^{-1} \!- \bH^{-1}\right) \bB.
	\end{equation}
\end{lemma}

\subsection{Declarative Layers for Blind PnP}
\label{sec:method_declarative}

In this section, we will demonstrate how the theory of bi-level optimization may be applied to the blind PnP problem.

\inlinesection{Weighted Blind PnP Layer}
This declarative layer operates on the $m \!\times\! n$ product set of 2D--3D correspondences, optimizing the lower-level objective function
\begin{equation}
\label{eqn:pnp_objective}
f(\bP, \br, \bt) = \sum_{i=1}^{m}\sum_{j=1}^{n} \bP_{ij} \left( 1 - \bbf_i\transpose \frac{\bR_{\br}\bp_j + \bt}{\|\bR_{\br}\bp_j + \bt\|} \right)
\end{equation}
over $\br \in \reals^3$ and $\bt \in \reals^3$, where $\bP$ is a fixed joint correspondence probability matrix with $\sum \bP_{ij} = 1$, a relaxation of the Boolean correspondence matrix $\bC$, and $\br$ is the angle-axis representation of the rotation $\bR_{\br}$ such that $\bR_{\br} = \exp{[\br]_{\times}}$ for the skew symmetric operator $[\cdot]_{\times}$.
The Rodrigues' rotation formula provides an efficient closed-form solution to this exponential map.
We use the angle-axis representation to automatically satisfy the constraints on $\bR$, that is, $\bR \in SO(3)$.
We minimize this nonlinear function using the native PyTorch L-BFGS optimizer \cite{byrd1995limited} to find $(\br^\star, \bt^\star)$ for a given joint probability matrix $\bP$.

Given optimal $(\br^\star, \bt^\star)$, corresponding to $\by$ in \lemref{lm:unconstrained}, we can compute the derivatives $\dd \br^\star (\bP)$ and $\dd \bt^\star (\bP)$ using \eqnref{eqn:unconstrained}.
Observe that the gradient computation is agnostic to the choice of optimization algorithm; we do not back-propagate through the L-BFGS iterations that were used to determine $(\br^\star, \bt^\star)$.
Instead, since the objective function is twice-differentiable, we only require that a (locally) optimal solution be found in order to compute the gradient in one step.
While an analytic solution for the gradient can be obtained, it is quite unwieldy.
In lieu of this, we use automatic differentiation to compute the necessary Jacobian and Hessian matrices.
To be clear, automatic differentiation is applied to the specification of the objective function, not the algorithmic steps used to optimize it, which is distinctly different from standard usage in deep learning.

Importantly, the $m \!\times\! n$ product set of correspondences is too large for existing differentiable PnP solvers, such as DLT \cite{dang2018eigendecomposition} and DSAC \cite{brachmann2017dsac}.
For $m\!=\!n\!=\!1000$, $99.9\%$ of the $mn$ possible correspondences are outliers.
Even with the weights $\bP$, outliers will dominate the solver.
We achieve robustness with top-$k$ RANSAC (see below) and nonlinear optimization.
In contrast, the non-robust linear estimate of DLT is unusably poor, and has severe numerical issues \cite{dang2018eigendecomposition}.
DSAC also fails in this case, because the probability of selecting an inlier hypothesis at random from the product set is vanishingly small.
Unlike our declarative layer, DSAC cannot differentiably select hypotheses from the top-$k$ subset.

\inlinesection{RANSAC}
While \lemref{lm:unconstrained} guarantees a local descent direction given the local minimizer $\by$, it is unlikely to be useful for the learning problem (or indeed the inference problem) if $\by$ is a bad estimate.
Hence it is helpful for $\by$ to be, on average, a good local optimum---preferably the global optimum.
Since the blind PnP objective function is non-convex with many local minima, a standard technique is to apply robust randomized global search such as RANSAC \cite{fischler1981random}.
The declarative framework gives us the opportunity to incorporate this non-differentiable algorithm into an end-to-end learning network.
We select a subset of $k=1.5\min\{m, n\}$ correspondences from the $mn$ possibilities, choosing those with the highest joint probability in $\bP$.
We then run RANSAC with the P3P algorithm \cite{gao2003complete} to find the inlier set, followed by the EPnP algorithm \cite{lepetit2009epnp} on all inliers to refine the estimate.
This robust estimate of the camera pose parameters is used to initialize the nonlinear weighted PnP optimization algorithm.

Since the final processing node in the declarative layer optimizes a twice-differentiable objective function, the non-differentiability of any intermediate computation is irrelevant to the gradient calculation.
Note that this procedure for robustly estimating the camera pose parameters has no analytic solution and involves a non-differentiable algorithm.
It would not be possible to use standard techniques such as explicit or automatic differentiation to obtain the gradient.

\inlinesection{Sinkhorn Layer}
We also define a declarative layer for feature matching in order to estimate the joint correspondence probability of the 2D--3D point pairs from a cost matrix $\bM \in \reals_{+}^{m \times n}$.
This is achieved by encapsulating the Sinkhorn algorithm \cite{sinkhorn1967diagonal} in a declarative layer, as has been previously demonstrated in the literature \cite{santacruz2018visual}.
The layer optimizes the lower-level objective function \cite{cuturi2013sinkhorn}
\begin{equation}
\label{eqn:sinkhorn_objective}
f(\bM, \bP) = \!\sum_{i=1}^m \sum_{j=1}^n \left(\bM_{ij} \bP_{ij} + \mu \bP_{ij} (\log{\bP_{ij}} - 1) \right)
\end{equation}
with respect to $\bP \in U(\br, \bc)$, where the transport polytope
\begin{equation}
\label{eqn:transport_polytope}
U(\br, \bc) = \{ \bP \in \reals_{+}^{m \times n} \mid \bP\ones^n = \br, \, \bP\transpose\ones^m = \bc \}
\end{equation}
is defined for the prior probability vectors $\br \in \reals_{+}^m$ and $\bc \in \reals_{+}^n$ with $\sum\br = 1$ and $\sum\bc = 1$, which represent the probability that any given 2D or 3D point has a valid match.
In this work, we use uniform priors $\br = \frac{1}{m}\ones$ and $\bc = \frac{1}{n}\ones$.

We run the highly-efficient Sinkhorn algorithm which optimizes this objective function, an entropy-regularized Wasserstein distance, in $O(m^2)$ \cite{cuturi2013sinkhorn}.
This is considerably more efficient than the Hungarian algorithm, which exactly optimizes the Wasserstein distance in $O(m^3)$, while also converging to the Wasserstein distance as $\mu \to 0$.
Given optimal $\bP^\star$, corresponding to $\by$ in \lemref{lm:linear_equality}, we can compute the derivative $\dd \bP^\star (\bM)$ using \eqnref{eqn:linear_equality}.
Unlike the PnP layer, we compute the derivative analytically to ensure memory efficiency.
To do so, we need to form the matrices $\bA$, $\bB$ and $\bH$ and perform the necessary inversions.
We defer the details to the supplementary material.

The benefits of enclosing this algorithm in a declarative layer include being able to run the algorithm to convergence, rather than fixing the number of iterations, and obviating the need for unrolling the algorithmic steps and maintaining the requisite computation graph, which saves a significant amount of memory.
Our implementation is much more memory efficient than that of Santa Cruz \etal \cite{santacruz2018visual}, reducing $O(m^2 n^2)$ memory requirements to $O(mn)$.
This allows much larger problems than were considered previously, such as $m=n=1000$ used in this work.
This was achieved by exploiting the block structure of the matrices $\bA$ and $\bH$ rather than storing them in full, and by computing the vector--Jacobian product rather than the Jacobian itself.

\subsection{Network Architecture}
\label{sec:method_arch}

Our network architecture is shown in \figref{fig:flowchart}.
First, we extract discriminative features from the 2D and 3D point-sets, aiming to recognise patterns in the data that are useful for establishing correspondences.
Next, we estimate the correspondence probability for every 2D--3D pair by computing the pairwise distance between features and solving an optimal transport problem.
Last, we optimize a weighted blind PnP objective function to obtain the (locally) optimal camera pose, given the data and estimated correspondence probability matrix.

\inlinesection{Feature Extraction}
\label{sec:method_features}
To extract discriminative features from the 2D and 3D point-sets, we directly use the point feature extraction model from Yi \etal \cite{yi2018learning}.
This model is a 12-layer ResNet \cite{he2016deep}, where each layer consists of a perceptron with 128 neurons per point, context normalization, batch normalization and a ReLU nonlinearity, with weights shared between points.
Context normalization is the mechanism for sharing information between points, by normalizing with respect to the mean $\bmu^l$ and standard deviation $\sigma^l$ of the feature vectors $\bz_i^l$ of every point at the $l$\textsuperscript{th} layer, and is given by $\text{CN}(\bz_i^l) = (\bz_i^l - \bmu^l) / \sigma^l$.

Before passing the data to the feature extraction networks, we do some initial processing. We convert the homogeneous bearing 3-vectors into inhomogeneous 2-vectors by dividing through by the z coordinate, since this requires fewer network parameters.
We also apply a learned $3\times3$ transformation matrix to the 3D points using the input transform from PointNet \cite{qi2017pointnet}, to align the points to a canonical orientation.
Finally, after obtaining the pointwise feature vectors, we apply $L_2$ normalization, which is helpful for the ensuing optimization procedure.

Hence the 2D feature extractor encodes a mapping $\Phi$ with parameters $\phi$ from the 2D bearing vector set $\cF$ to the feature vector set $\cZ_\cF$, given by $\cZ_\cF = \Phi_\phi (\cF)$ with $\cZ_\cF = \{\bz_{\bbf_i}\}_{i=1}^m$ and $\bz_{\bbf_i} \in \reals^{128}$.
The 3D feature extractor encodes a similar mapping, given by $\cZ_\cP = \Psi_\psi (\cP)$ with $\cZ_\cP = \{\bz_{\bp_i}\}_{i=1}^n$ and $\bz_{\bp_i} \in \reals^{128}$.

\inlinesection{Correspondence Probability}
To estimate the probability that a 2D--3D point pair is an inlier correspondence, we compute the pairwise distances between the feature vector sets $\cZ_\cF$ and $\cZ_\cP$ and then solve an optimal transport problem,
as was shown to be effective in concurrent work \cite{liu2019deep}.
The elements of the pairwise $L_2$ distance matrix $\bM \in \reals_{+}^{m \times n}$ are computed as $\bM_{ij} = \| \bz_{\bbf_i} - \bz_{\bp_j} \|_2$.
We then solve the regularized transport problem \cite{cuturi2013sinkhorn} using the Sinkhorn algorithm \cite{sinkhorn1967diagonal} to obtain a joint probability matrix $\bP$.
The advantage of this approach is that it considers the entirety of $\bM$ when estimating the probability $\bP_{ij}$, in order to resolve correspondence ambiguities.
Finding a jointly optimal solution is critical if the learning goal is to approach the \emph{one-to-one} correspondence matrix $\bC$ up to scale.
This optimization problem was outlined in \secref{sec:method_declarative}, where we showed that the gradient computation can be decoupled from the Sinkhorn algorithm using implicit differentiation.

\inlinesection{Blind PnP Optimization}
\label{sec:method_pnp}
Given the joint correspondence probability matrix $\bP$, we can now optimize the weighted nonlinear blind PnP objective function \eqnref{eqn:pnp_objective} to obtain the optimal camera pose $(\bR, \bt)$ for that set of correspondence probabilities.
The optimization problem was outlined in \secref{sec:method_declarative}, where we showed how to find a locally-optimal camera pose using the L-BFGS algorithm, how to ensure it is a good local optimum (on average) using RANSAC, and how to back-propagate through the layer.
Hence we have a fully-differentiable way to generate camera pose parameters that are likely to be near the global optimum of the non-convex objective function.
At test time, we can either take the network output or the RANSAC output computed within the network; both are evaluated in the experiments.

\subsection{Learning From Pose-Labelled Data}
\label{sec:method_learning}

\inlinesection{Loss Functions}
We use two loss functions, one for each component of the coupled problem.
The first is a correspondence loss $L_\text{c}$ to bring the estimated correspondence matrix $\bP$ closer to the ground-truth.
The second is a pose loss $L_\text{p}$ to encourage the network to generate correspondence matrices that are amenable to our PnP solver.
Note that these are distinct, albeit complementary, aims.
While a perfect correspondence matrix would generate an accurate camera pose, this is not achievable in practice.
Instead, there is a family of correspondence matrices for a fixed suboptimal value of $L_\text{c}$, which will differ considerably in their suitability for the weighted PnP solver.

The correspondence loss $L_\text{c}$ arises directly from the problem formulation \eqnref{eqn:optimisation_problem} given the ground-truth rotation $\bR_\text{gt}$ and translation $\bt_\text{gt}$, yielding
\begin{equation}
\label{eqn:loss_correspondences}
L_\text{c} = \sum_{i}^{m}\sum_{j}^{n} \bP_{ij} \left(1 - 2 \bigind{ \angle \left(\bbf_i, \bR_\text{gt}\bp_j + \bt_\text{gt} \right) \leqslant \theta} \right)
\end{equation}
where \ind{\!\cdot\!} is an Iverson bracket (indicator function), and $\theta$ is the angular inlier threshold.
The loss is bounded, since $\sum \bP_{ij}=1$ and so $L_\text{c} \in [-1,1)$, and has the interpretation of maximizing the probability of the inlier correspondences and minimizing the probability of the outlier correspondences, since $\bP$ is a joint probability matrix.
If the ground-truth correspondence matrix $\bC$ is available, this can be used instead of the indicator function.
It is also possible to use the (less robust) reprojection error \eqnref{eqn:reprojection_error}, however we found no advantage to this.

The camera pose loss $L_\text{p}$ uses standard error measures on rotations and translations, and is given by
\begin{align}
L_\text{p} &= L_\text{r} + L_\text{t}\label{eqn:loss_pose}\\
L_\text{r} &= \angle \big( \bR, \bR_\text{gt} \big) = \arccos{\tfrac{1}{2} \big(\trace{\bR_\text{gt}\transpose\bR} - 1 \big)}\label{eqn:loss_rotation}\\
L_\text{t} &= \| \bt - \bt_\text{gt}\|_2\label{eqn:loss_translation}
\end{align}
This loss is not bounded, since $L_\text{r} \in [0, \pi]$ and $L_\text{t} \in [0, \infty)$.
The argument of $\arccos$ is clamped to between $\pm (1-\epsilon)$ for $\epsilon = 10^{-7}$ to prevent an infinite gradient at $0^\circ$ and $180^\circ$.
Finally, the total loss is given by
\begin{equation}
\label{eqn:loss_total}
L = L_\text{c} + \gamma_\text{p} L_\text{p}
\end{equation}
where $\gamma_\text{p}$ is a hyperparameter that controls the relative influence of $L_\text{p}$.

\inlinesection{Learning Strategy}
We train the network implemented in PyTorch using the Adam optimizer \cite{kingma2015adam} with a learning rate of $10^{-5}$ and otherwise default parameters. We use a batch size of $16$ and train for 120 epochs
(to convergence) with the correspondence loss only ($\gamma_\text{p} = 0$), followed by 20--80 epochs with the pose loss as well ($\gamma_\text{p} = 1$).
This reflects the intuition that the pose loss is more meaningful once the correspondence probability matrix $\bP$ has useful information, having reduced the correspondence search space.

\inlinesection{Implementation Details}
For the Sinkhorn algorithm, the entropy parameter $\mu$ was set to $0.1$; for RANSAC, the inlier reprojection error was set to $0.01$ and the maximum number of iterations was set to $1000$; and for the L-BFGS solver, the line search function was set to strong Wolfe, the maximum number of iterations varied with the number of points to standardize the batch runtime, and the gradient norms were clipped to $100$.
Ground-truth correspondences were used in training instead of specifying inlier threshold $\theta$.
All experiments were run on a single Titan V GPU, and the PyTorch code, including modular Sinkhorn and weighted PnP layers, is available at \href{https://github.com/dylan-campbell/bpnpnet}{github.com/dylan-campbell/bpnpnet}.

\section{Results}
\label{sec:results}

Our blind PnP network, named BPnPNet, is evaluated with respect to the baseline algorithms SoftPOSIT \cite{david2004softposit}, RANSAC \cite{fischler1981random}, and GOSMA \cite{campbell2019alignment} on synthetic and real data.
These are state-of-the-art representative examples of a local blind PnP solver (SoftPOSIT), a global solver (RANSAC), and a globally-optimal solver (GOSMA).
For RANSAC, we randomly sample 2D--3D correspondences and use a minimal P$3$P solver \cite{kneip2011novel}.\footnote{The probability of choosing a minimal set of 4 true 2D--3D correspondences from the size $mn$ set of all correspondences without replacement is $\prod_{i=0}^{3} \frac{m-i}{(m-i)(n-i)} \approx 10^{-12}$ for $m=n=1000$ and no outliers. The number of RANSAC iterations required to achieve 90\% confidence is thus $\log(1 - 0.9) / \log(1 - 10^{-12}) \approx 2.3\times10^{12}$.}
For SoftPOSIT, we provide an initialization using ground-truth pose information, since it is a local solver and therefore requires a good pose prior.
The algorithms were stopped early if their runtime for a single point-set pair exceeded $30$s, returning the best pose found so far.
This ensured that evaluation time was bounded at four days per algorithm on the datasets tested.
Globally-optimal algorithms often exceed this limit, but it is infeasible to evaluate them to convergence on large datasets.

We use the synthetic ModelNet40 dataset \cite{wu20153d} and the real-world MegaDepth dataset \cite{li2018megadepth} for evaluation.
The former is a CAD mesh model dataset, while the latter is a multi-view photo dataset with COLMAP \cite{schoenberger2016sfm} reconstructions providing the 2D and 3D point-sets.
MegaDepth has highly diverse scenes, camera poses, and point distributions.
We report quartiles for rotation error (in degrees), translation error, and reprojection error (in degrees), according to \eqnref{eqn:loss_rotation}, \eqnref{eqn:loss_translation}, and \eqnref{eqn:reprojection_error} respectively.
We denote the first, second (median) and third quartiles as Q1, Q2 and Q3.
We also report average runtime for inference (in seconds) and recall at a particular error threshold (as a percentage), that is, the percentage of poses with an error less than that threshold.

\subsection{Synthetic Data Experiments}
\label{sec:results_synthetic}

In this section, we evaluate our network on the synthetic ModelNet40 dataset \cite{wu20153d} and conduct ablation studies.
To generate the synthetic data from the mesh models, we uniformly sampled $1000$ 3D points from each model and generated virtual cameras by drawing Euler rotation angles uniformly from $[0, \pi/4]$ and translations from $[-0.5, 0.5]$, with an offset of 4.5 along the $z$ axis.
The points were projected to a $640\times480$ virtual image with a focal length of $800$ and normal noise with $\sigma = 2$ pixels was applied to the 2D points.
In this way, we generated training and testing sets of $40000$  and $2468$ 2D--3D point-set pairs respectively, each from the standard train and test splits of ModelNet40.
SoftPOSIT was initialized using the mean ground-truth Euler angles and translation, corresponding to a median initial rotation error of $21.5^\circ$ and translation error of $0.49$.

\begin{table}[!t]\centering
	\caption{
		Results on the ModelNet40 \cite{wu20153d} test set.
		We report quartiles for rotation error  ($^{\circ}$), translation error and reprojection error  ($^{\circ}$), and the mean runtime T (s).
		Note that Ours $L_\text{c}R$ is a standard RANSAC baseline: deep 2D--3D feature matching followed by P3P-RANSAC.
		$^\dagger\!$Algorithms were run for a maximum of 30s.
	}
	\label{tab:modelnet}
	\newcolumntype{C}{>{\centering\arraybackslash}X}
\renewcommand{\arraystretch}{0.88} 
\begin{tabularx}{\linewidth}{@{}l C C C C C C C C C c@{}}
	\toprule
	& \multicolumn{3}{c}{Rotation Error} & \multicolumn{3}{c}{Translation Error} & \multicolumn{3}{c}{Reproj. Error} & T\\
	\cmidrule(l{3pt}r{3pt}){2-4}
	\cmidrule(l{3pt}r{3pt}){5-7}
	\cmidrule(l{3pt}r{3pt}){8-10}
	\cmidrule(l{0pt}r{0pt}){11-11}
	Method & Q1 & Q2 & Q3 & Q1 & Q2 & Q3 & Q1 & Q2 & Q3 & $\bar{x}$\\
	\midrule
	SoftPOSIT \cite{david2004softposit} & 16.1 & 21.8 & 28.0 & 0.33 & 0.49 & 0.72 & 2.82 & 3.98 & 5.21 & 27$^\dagger$\\
	RANSAC \cite{fischler1981random} & 90.8 & 139 & 165 & 0.43 & 1.15 & 3.08 & 4.22 & 5.87 & 8.06 & 30$^\dagger$\\
	GOSMA \cite{campbell2019alignment} & 10.1 & 22.1 & 52.0 & 0.25 & 0.46 & 0.75 & 1.04 & 1.62 & 3.11 & 30$^\dagger$\\
	\midrule
	Ours $L_\text{c}$ & 6.08 & 11.3 & 18.3 & 0.34 & 0.52 & 0.81 & 0.56 & 0.86 & 1.31 & \textbf{0.1}\\
	Ours $L_\text{c} L_\text{p}$ & 4.88 & 9.66 & 16.0 & \textbf{0.04} & \textbf{0.08} & \textbf{0.15} & 0.36 & 0.61 & 1.03 & \textbf{0.1}\\
	Ours $L_\text{c}R$ & 5.49 & 11.7 & 20.0 & \textbf{0.04} & 0.09 & 0.20 & 0.37 & 0.70 & 1.25 & \textbf{0.1}\\
	Ours $L_\text{c} L_\text{p} R$ & \textbf{3.33} & \textbf{8.09} & \textbf{15.8} & \textbf{0.04} & \textbf{0.08} & 0.16 & \textbf{0.28} & \textbf{0.52} & \textbf{1.01} & \textbf{0.1}\\
	\bottomrule
\end{tabularx}%

\end{table}
\begin{figure}[!t]\centering
	\includegraphics[width=\textwidth]{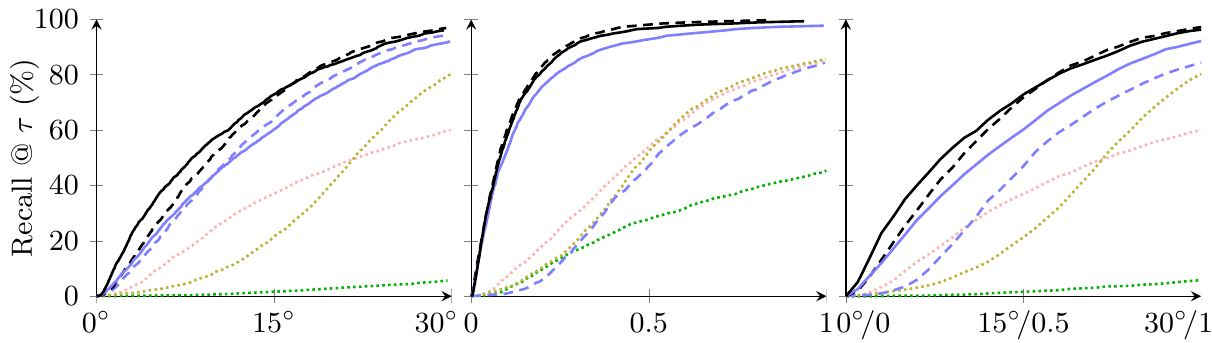}\vspace{-6pt}
	\includegraphics[width=\textwidth, trim=0 8 0 0, clip]{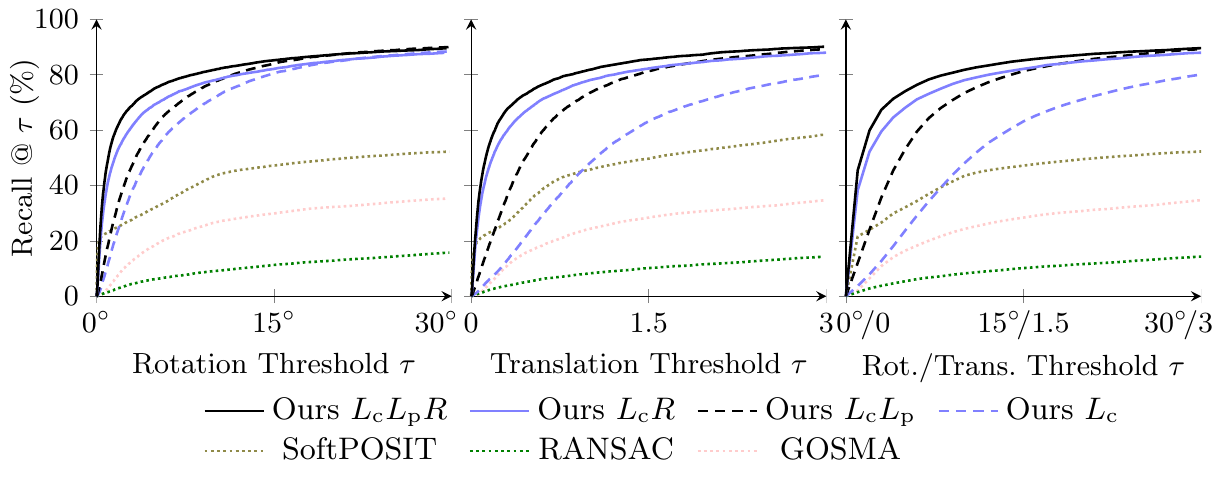}
	\caption{
		Recall on the ModelNet40 (top) and MegaDepth (bottom) test sets, with respect to an error threshold $\tau$.
		$R$ denotes using the RANSAC estimate.
	}
	\label{fig:results}
\end{figure}

The results on the ModelNet40 test set are shown in \tabref{tab:modelnet} and \figref{fig:results}.
They demonstrate that our network obtains significantly better camera pose results than state-of-the-art local, global and globally-optimal algorithms.
The results also include our ablation study, where we compare our model's camera pose output (Ours $L_\text{c} L_\text{p}$) with a variant of our model that is learnt without the pose loss (Ours $L_\text{c}$).
In all cases, the pose loss improves the results significantly, especially the translation errors.
In particular, the recall for rotation errors less than $15^\circ$ and translation errors less than $0.5$ is $72\%$, an improvement of $25\%$ over using the $L_\text{c}$ loss only.
We also compare our model's output with the RANSAC pose computed within our network, denoted by $R$ in the results and ($\bR_0, \bt_0$) in \figref{fig:flowchart}.
The robust RANSAC estimate tends to be more accurate than the model's final output since it is more resistant to errors in the correspondence probability matrix, and so should be used in any applications.
Our method is also at least two orders of magnitude faster than the other methods, taking $0.12$s on average.
Note that Ours $L_\text{c}R$ is an example of the standard RANSAC baseline of deep 2D--3D feature matching followed by P3P-RANSAC, without end-to-end training.
Finally, a small reduction of only $2^\circ/0.01$ on the median statistics is observed with weaker pose-only supervision (see supplementary material).

\subsection{Real Data Experiments}
\label{sec:results_real}

\begin{figure}[!t]\centering
	\includegraphics[width=0.37\textwidth, trim=4pt 4pt 4pt 110pt, clip]{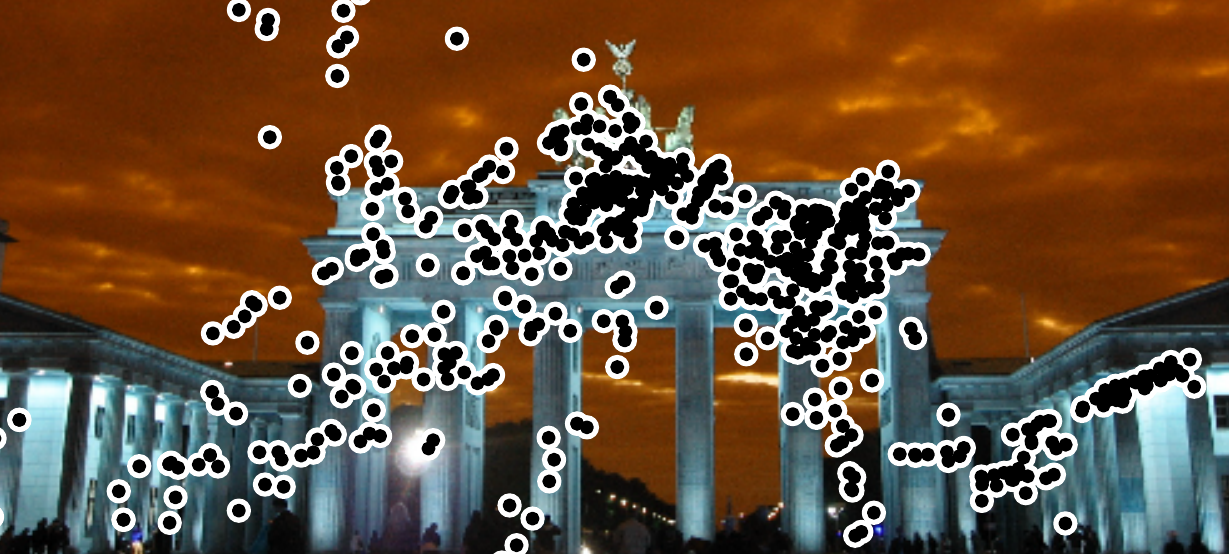}\hfill%
	\includegraphics[width=0.23\textwidth, trim=65pt 40pt 5pt 14pt, clip]{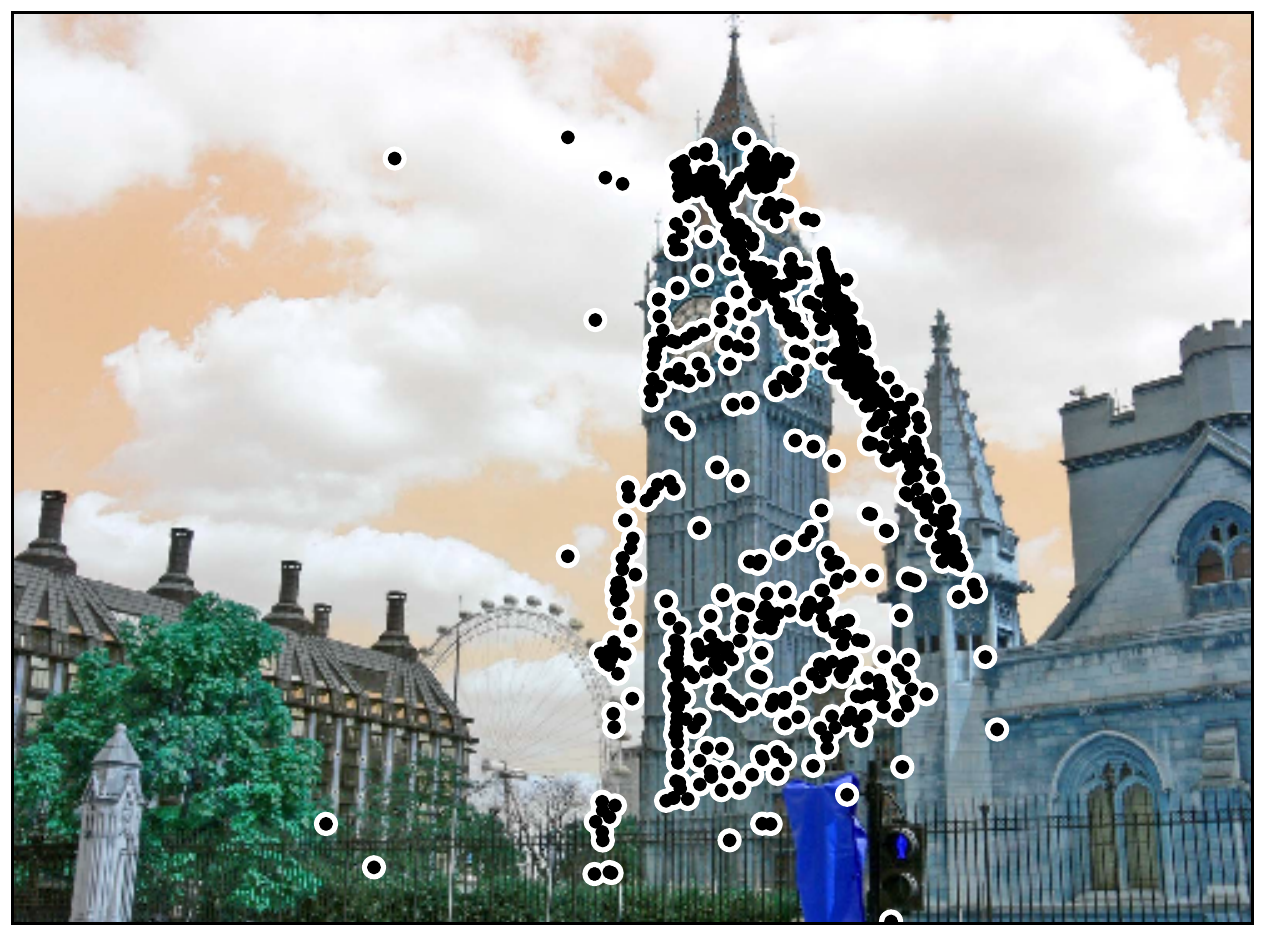}\hfill%
	\includegraphics[width=0.37\textwidth, trim=4pt 64pt 4pt 50pt, clip]{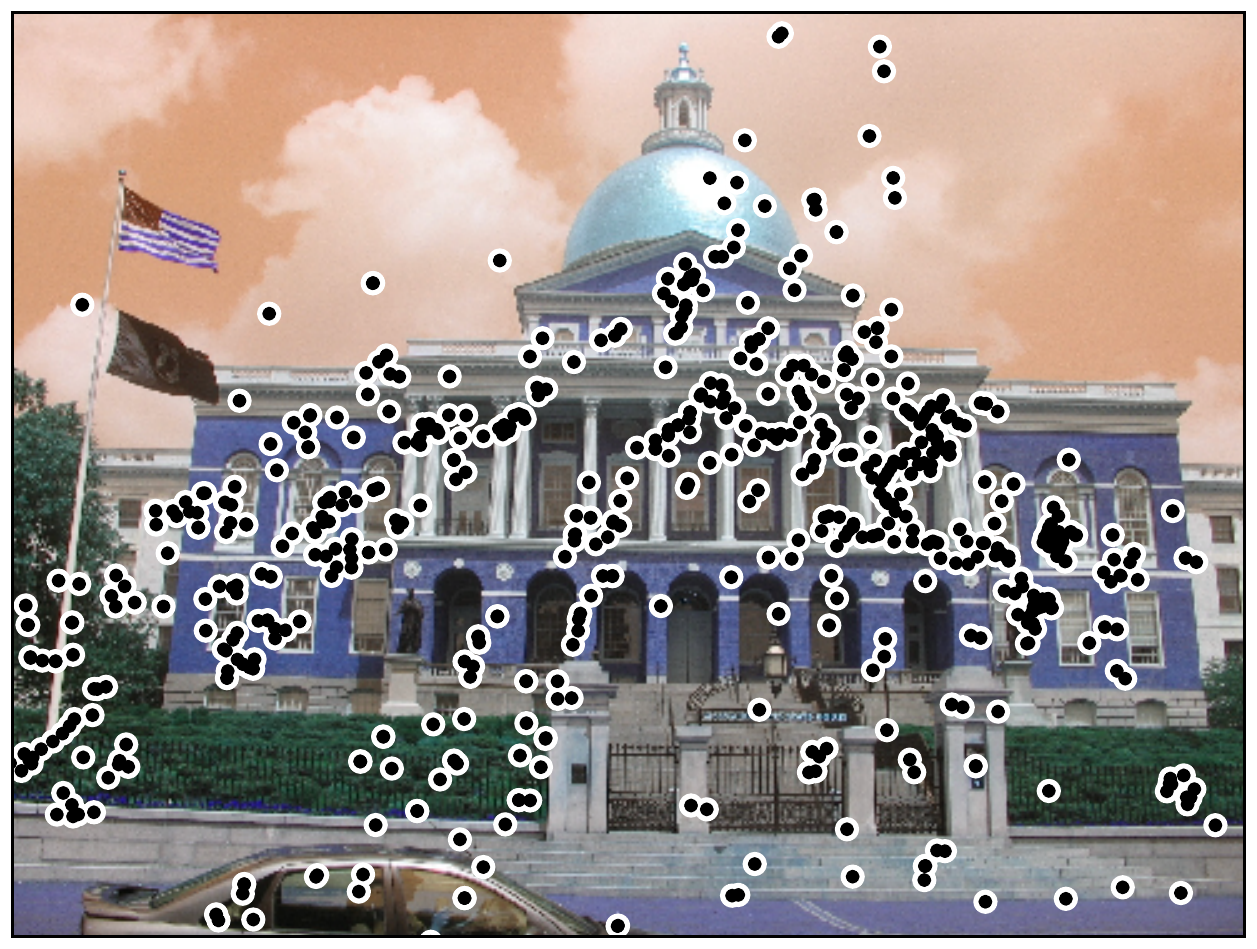}\vspace{-1pt}
	\includegraphics[width=0.37\textwidth, trim=4pt 4pt 4pt 110pt, clip]{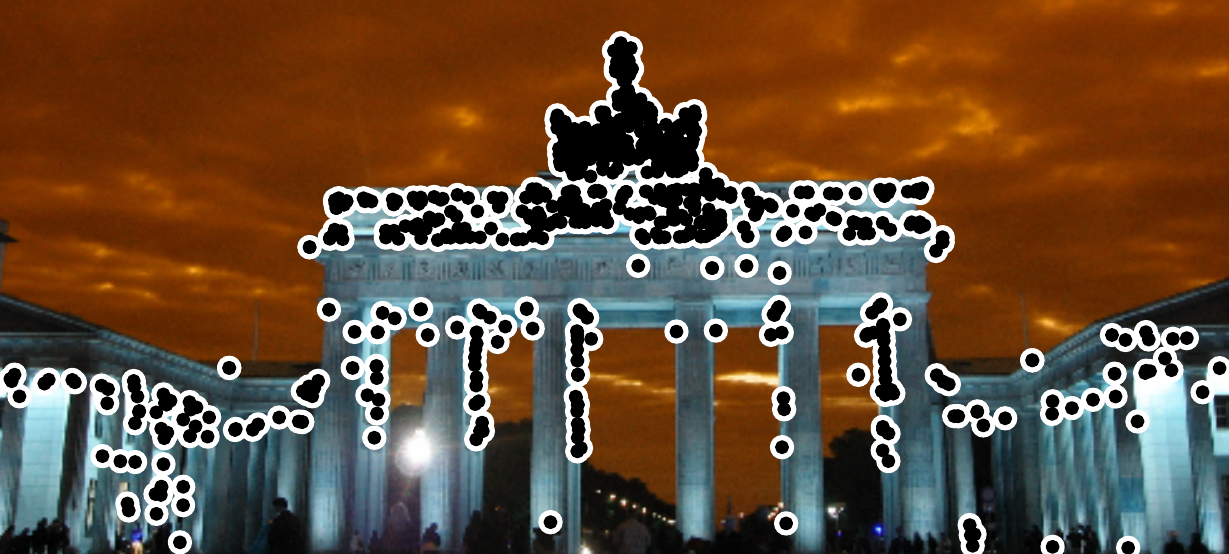}\hfill%
	\includegraphics[width=0.23\textwidth, trim=65pt 40pt 5pt 14pt, clip]{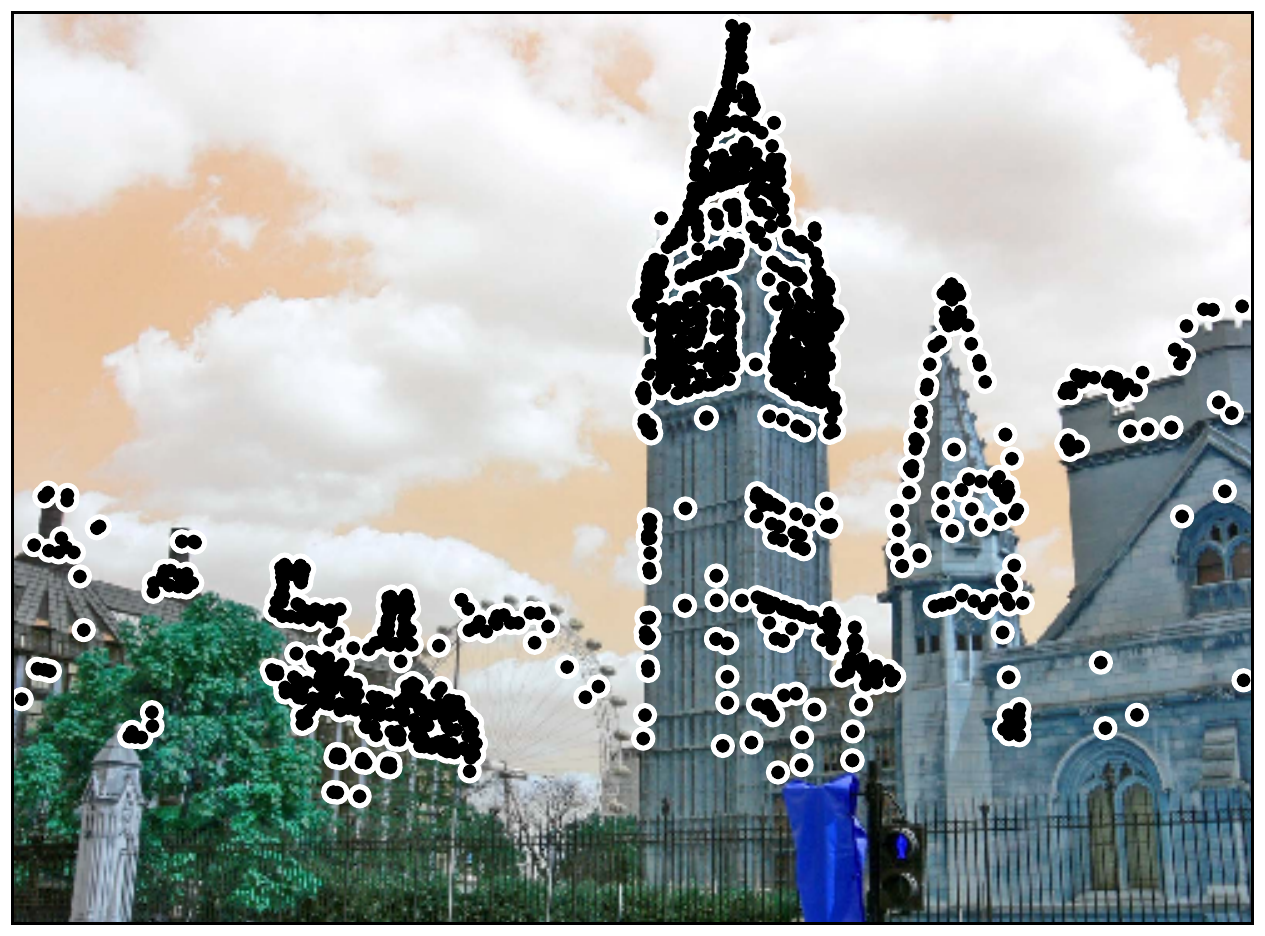}\hfill%
	\includegraphics[width=0.37\textwidth, trim=4pt 64pt 4pt 50pt, clip]{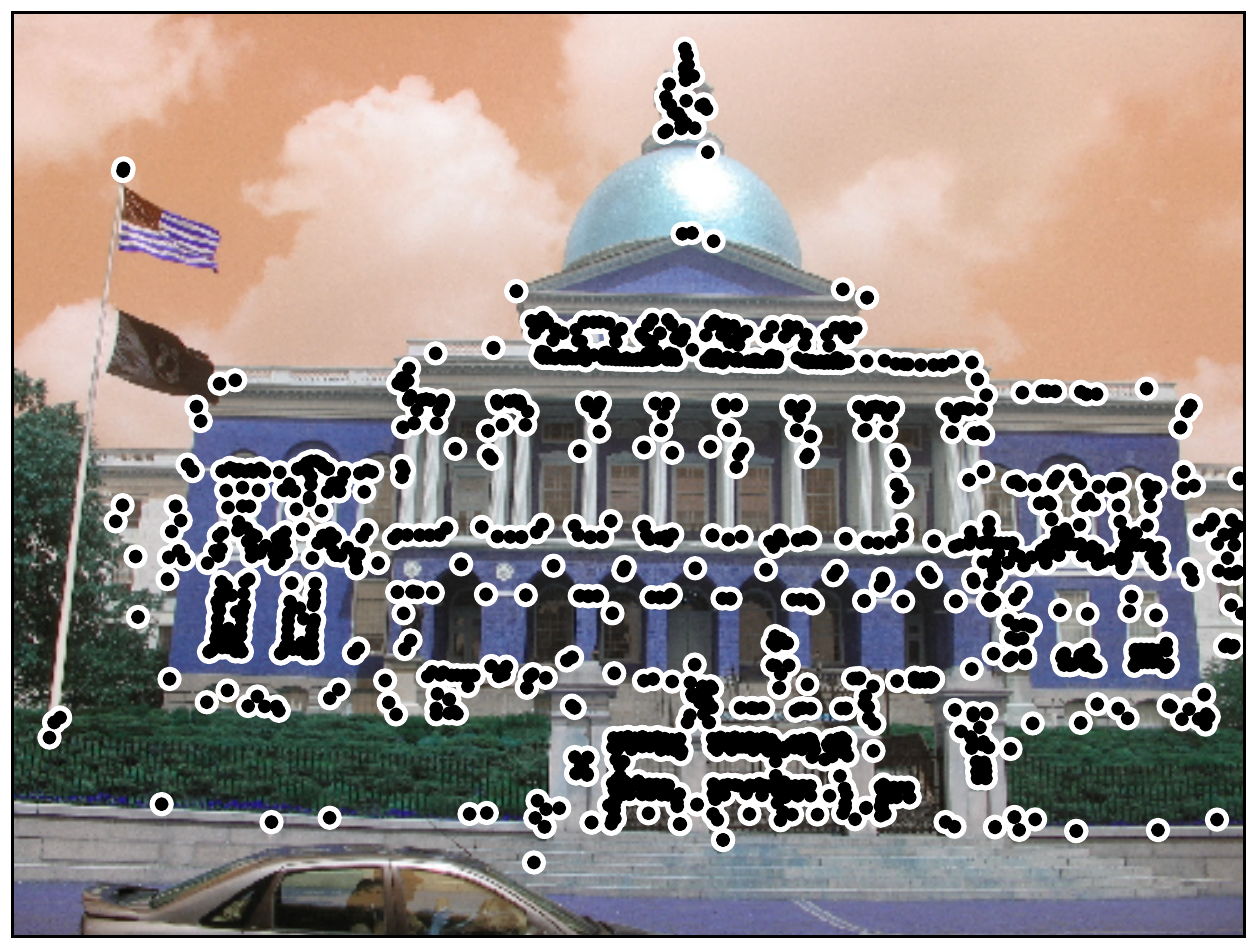}%
	\caption{
		Qualitative results for the MegaDepth dataset: 3D point-sets projected onto the image using the camera pose found by GOSMA \cite{campbell2019alignment} (top) and our method (bottom).
	}
	\label{fig:results_megadepth_qualitative}
\end{figure}

\begin{table}[!t]\centering\footnotesize
	\caption{
		Results on the MegaDepth \cite{li2018megadepth} test set.
		We report quartiles for rotation error  ($^{\circ}$), translation error and reprojection error  ($^{\circ}$), and the mean runtime T (s).
		$^\dagger\!$Algorithms were run for a maximum of 30s.
	}
	\label{tab:megadepth}
	\newcolumntype{C}{>{\centering\arraybackslash}X}
\renewcommand*{\arraystretch}{0.70} 
\setlength{\tabcolsep}{0pt} 
\begin{tabularx}{\linewidth}{@{}l C C C C C C C C C c@{}}
	\toprule
	& \multicolumn{3}{c}{Rotation Error} & \multicolumn{3}{c}{Translation Error} & \multicolumn{3}{c}{Reproj. Error} & T\\
	\cmidrule(l{3pt}r{3pt}){2-4}
	\cmidrule(l{3pt}r{3pt}){5-7}
	\cmidrule(l{3pt}r{3pt}){8-10}
	\cmidrule(l{0pt}r{0pt}){11-11}
	Method & Q1 & Q2 & Q3 & Q1 & Q2 & Q3 & Q1 & Q2 & Q3 & $\bar{x}$\\
	\midrule
	SoftPOSIT \cite{david2004softposit} & 1.81 & 21.4 & 165 & 0.24 & 1.53 & 6.10 & 0.92 & 7.85 & 24.1 & 18$^\dagger$\\
	RANSAC \cite{fischler1981random} & 66.6 & 122 & 155 & 6.80 & 15.2 & 28.2 & 4.45 & 8.77 & 13.3 & 30$^\dagger$\\
	GOSMA \cite{campbell2019alignment} & 8.69 & 86.8 & 145 & 1.07 & 5.67 & 9.34 & 1.30 & 13.7 & 37.1 & 30$^\dagger$\\
	\midrule
	Ours $L_\text{c}$ & 1.91 & 4.47 & 11.4 & 0.52 & 1.05 & 2.34 & 0.54 & 1.12 & 2.81 & \textbf{0.2}\\
	Ours $L_\text{c} L_\text{p}$ & 1.32 & 3.31 & 8.84 & 0.21 & 0.46 & 1.08 & 0.21 & 0.53 & 1.64 & \textbf{0.2}\\
	Ours $L_\text{c}R$ & 0.44 & 1.55 & 7.70 & 0.05 & 0.18 & 0.80 & 0.06 & 0.16 & 1.27 & \textbf{0.2}\\
	Ours $L_\text{c} L_\text{p} R$ & \textbf{0.34} & \textbf{1.00} & \textbf{4.88} & \textbf{0.04} & \textbf{0.12} & \textbf{0.53} & \textbf{0.06} & \textbf{0.12} & \textbf{0.74} & \textbf{0.2}\\
	\bottomrule
\end{tabularx}%

\end{table}

In this section, we evaluate our network on the real-world MegaDepth dataset \cite{li2018megadepth}.
To generate the splits, we randomly selected landmarks and obtained training and testing sets of $40828$  and $10795$ 2D--3D point-set pairs respectively.
The landmarks in the different splits do not overlap.
That is, the experiment measures how well the network \emph{generalizes} to unseen locations, often in different countries.
There is an extremely wide range of point-set sizes in this dataset, with the number of points in the training set varying by four orders of magnitude from $4$ to $14742$, reflecting the variability of real-world structure-from-motion data.
SoftPOSIT was initialized using the ground-truth, with an angular perturbation drawn uniformly from $[-10, 10]$ degrees about a random axis, and a translation perturbation drawn uniformly from $[-0.5, 0.5]$ in a random direction.
As such, it cannot be compared to the other methods directly.

The results on the MegaDepth test set are shown in \tabref{tab:megadepth} and \figref{fig:results}, with qualitative results given in \figref{fig:results_megadepth_qualitative}.
Our approach outperforms the other algorithms by a significant margin, notably doing better than a local optimization algorithm initialized very close to the ground-truth.
GOSMA performs poorly on this dataset, despite being optimal, since it rarely converges within the $30$s evaluation limit (often taking minutes).
As with the synthetic data, the pose loss improves the results, especially the translation errors.
In particular, the recall for rotation errors less than $10^\circ$ and translation errors less than $1$ is $73\%$, $25\%$ better than without the loss.
With the RANSAC estimate, the recall further improves to $82\%$.
Additional results, including outlier analysis, failure cases, and another feature matching approach, are provided in the supplementary material.

\section{Conclusion}
\label{sec:conclusion}

In this paper, we have proposed the first fully end-to-end trainable network for solving the blind PnP problem.
The key insight is that we can back-propagate through a geometric optimization algorithm using the technique of implicit differentiation.
This allows us to compute a gradient even when the declarative layer involves non-differentiable RANSAC search and L-BFGS optimization of a nonlinear geometric objective.
For such a layer, unrolling the algorithmic steps and computing the gradient with automatic differentiation is not possible, and would not be advisable even if it were due to the memory and computational requirements.
Furthermore, we show that our method outperforms state-of-the-art geometric blind PnP solvers by a considerable margin when pose-labelled training data is available.
Promisingly, our declarative approach admits the possibility of an unsupervised reprojection error loss, which may be used to fine-tune our pre-trained model to test scene data without the need for ground-truth labels.


%
%
\bibliographystyle{splncs04}

\begin{thebibliography}{10}
\providecommand{\url}[1]{\texttt{#1}}
\providecommand{\urlprefix}{URL }
\providecommand{\doi}[1]{https://doi.org/#1}

\bibitem{agrawal2019differentiable}
Agrawal, A., Amos, B., Barratt, S., Boyd, S., Diamond, S., Kolter, Z.:
  Differentiable convex optimization layers. In: Wallach, H., Larochelle, H.,
  Beygelzimer, A., d'Alch\'{e} Buc, F., Fox, E., Garnett, R. (eds.) Advances in
  Neural Information Processing Systems 32 (NIPS 2019). pp. 9562--9574. Curran
  Associates, Inc. (2019)

\bibitem{agrawal2019differentiating}
Agrawal, A., Barratt, S., Boyd, S., Busseti, E., Moursi, W.: Differentiating
  through a cone program. Journal of Applied and Numerical Optimization
  \textbf{1}(2),  107--115 (2019)

\bibitem{amos2017optnet}
Amos, B., Kolter, J.Z.: {O}pt{N}et: Differentiable optimization as a layer in
  neural networks. In: Precup, D., Teh, Y.W. (eds.) Proceedings of the 34th
  International Conference on Machine Learning. Proceedings of Machine Learning
  Research, vol.~70, pp. 136--145. PMLR, International Convention Centre,
  Sydney, Australia (06--11 Aug 2017)

\bibitem{baka2014oriented}
Baka, N., Metz, C., Schultz, C.J., van Geuns, R.J., Niessen, W.J., van Walsum,
  T.: Oriented {G}aussian mixture models for nonrigid {2D/3D} coronary artery
  registration. IEEE Transactions on Medical Imaging  \textbf{33}(5),
  1023--1034 (May 2014). \doi{10.1109/TMI.2014.2300117}

\bibitem{bard1998practical}
Bard, J.F.: Practical Bilevel Optimization: Algorithms and Applications. Kluwer
  Academic Press (1998)

\bibitem{besl1992method}
Besl, P.J., McKay, N.D.: A method for registration of {3-D} shapes. {IEEE}
  Trans.~on Pattern Analysis and Machine Intelligence ({PAMI})  \textbf{14}(2),
   239--256 (1992)

\bibitem{brachmann2017dsac}
Brachmann, E., Krull, A., Nowozin, S., Shotton, J., Michel, F., Gumhold, S.,
  Rother, C.: {DSAC} -- {D}ifferentiable {RANSAC} for camera localization. In:
  Proceedings of the 2017 Conference on Computer Vision and Pattern
  Recognition. pp. 2492--2500. IEEE Computer Society (Jul 2017).
  \doi{10.1109/CVPR.2017.267}

\bibitem{brachmann2018learning}
Brachmann, E., Rother, C.: Learning less is more - {6D} camera localization via
  {3D} surface regression. In: Proceedings of the 2018 Conference on Computer
  Vision and Pattern Recognition. pp. 4654--4662. IEEE Computer Society (2018)

\bibitem{brown2015globally}
Brown, M., Windridge, D., Guillemaut, J.Y.: Globally optimal {2D-3D}
  registration from points or lines without correspondences. In: Proceedings of
  the 2015 International Conference on Computer Vision. pp. 2111--2119 (Dec
  2015)

\bibitem{byrd1995limited}
Byrd, R.H., Lu, P., Nocedal, J., Zhu, C.: A limited memory algorithm for bound
  constrained optimization. SIAM Journal on Scientific Computing
  \textbf{16}(5),  1190--1208 (1995)

\bibitem{campbell2018globally}
Campbell, D., Petersson, L., Kneip, L., Li, H.: Globally-optimal inlier set
  maximisation for camera pose and correspondence estimation. {IEEE} Trans.~on
  Pattern Analysis and Machine Intelligence ({PAMI})  \textbf{42}(2),  328--342
  (Feb 2020). \doi{10.1109/TPAMI.2018.2848650}

\bibitem{campbell2019alignment}
Campbell, D., Petersson, L., Kneip, L., Li, H., Gould, S.: The alignment of the
  spheres: Globally-optimal spherical mixture alignment for camera pose
  estimation. In: Proceedings of the 2019 Conference on Computer Vision and
  Pattern Recognition. pp. 11796--11806. IEEE Computer Society (2019)

\bibitem{chen2020end}
Chen, B., Parra, A., Cao, J., Li, N., Chin, T.J.: End-to-end learnable
  geometric vision by backpropagating {PnP} optimization. In: Proc.~of the
  {IEEE} Conference on Computer Vision and Pattern Recognition ({CVPR}). IEEE
  Computer Society (Jun 2020)

\bibitem{cherian2017generalized}
Cherian, A., Fernando, B., Harandi, M., Gould, S.: Generalized rank pooling for
  action recognition. In: Proc.~of the {IEEE} Conference on Computer Vision and
  Pattern Recognition ({CVPR}). IEEE Computer Society (2017)

\bibitem{cuturi2013sinkhorn}
Cuturi, M.: Sinkhorn distances: Lightspeed computation of optimal transport.
  In: Burges, C.J.C., Bottou, L., Welling, M., Ghahramani, Z., Weinberger, K.Q.
  (eds.) Advances in Neural Information Processing Systems ({NeurIPS}). pp.
  2292--2300. Curran Associates Inc. (2013)

\bibitem{dang2018eigendecomposition}
Dang, Z., Moo~Yi, K., Hu, Y., Wang, F., Fua, P., Salzmann, M.:
  Eigendecomposition-free training of deep networks with zero eigenvalue-based
  losses. In: Ferrari, V., Hebert, M., Sminchisescu, C., Weiss, Y. (eds.)
  Proc.~of the European Conference on Computer Vision ({ECCV}). pp. 768--783.
  Springer (2018)

\bibitem{david2004softposit}
David, P., Dementhon, D., Duraiswami, R., Samet, H.: {SoftPOSIT:} simultaneous
  pose and correspondence determination. International Journal of Computer
  Vision ({IJCV})  \textbf{59}(3),  259--284 (2004)

\bibitem{dontchev2014implicit}
Dontchev, A.L., Rockafellar, R.T.: Implicit Functions and Solution Mappings: A
  View from Variational Analysis. Springer-Verlag, 2nd edn. (2014)

\bibitem{enqvist2008robust}
Enqvist, O., Kahl, F.: Robust optimal pose estimation. In: Forsyth, D., Torr,
  P., Zisserman, A. (eds.) Proceedings of the 2008 European Conference on
  Computer Vision. pp. 141--153. Springer (Oct 2008)

\bibitem{fathy2018hierarchical}
Fathy, M.E., Tran, Q.H., Zeeshan~Zia, M., Vernaza, P., Chandraker, M.:
  Hierarchical metric learning and matching for {2D} and {3D} geometric
  correspondences. In: Ferrari, V., Hebert, M., Sminchisescu, C., Weiss, Y.
  (eds.) Proc.~of the European Conference on Computer Vision ({ECCV}). pp.
  803--819. Springer (2018)

\bibitem{fernando2016learning}
Fernando, B., Gould, S.: Learning end-to-end video classification with
  rank-pooling. In: Balcan, M.F., Weinberger, K.Q. (eds.) Proc.~of the
  International Conference on Machine Learning ({ICML}). PMLR (2016)

\bibitem{fernando2017discriminatively}
Fernando, B., Gould, S.: Discriminatively learned hierarchical rank pooling
  networks. In: International Journal of Computer Vision ({IJCV}). vol.~124,
  pp. 335--355 (Sep 2017)

\bibitem{fischler1981random}
Fischler, M.A., Bolles, R.C.: Random sample consensus: a paradigm for model
  fitting with applications to image analysis and automated cartography.
  Communications of the ACM  \textbf{24}(6),  381--395 (1981)

\bibitem{gao2003complete}
Gao, X.S., Hou, X.R., Tang, J., Cheng, H.F.: Complete solution classification
  for the perspective-three-point problem. IEEE Transactions on Pattern
  Analysis and Machine Intelligence  \textbf{25}(8),  930--943 (Aug 2003)

\bibitem{gould2019deep}
Gould, S., Hartley, R., Campbell, D.: Deep declarative networks: A new hope.
  Tech. rep., Australian National University (arXiv:1909.04866) (Sep 2019)

\bibitem{he2016deep}
He, K., Zhang, X., Ren, S., Sun, J.: Deep residual learning for image
  recognition. In: Proc.~of the {IEEE} Conference on Computer Vision and
  Pattern Recognition ({CVPR}). IEEE Computer Society (2016)

\bibitem{kendall2017geometric}
Kendall, A., Cipolla, R.: Geometric loss functions for camera pose regression
  with deep learning. In: Proceedings of the 2017 Conference on Computer Vision
  and Pattern Recognition. pp. 6555--6564. IEEE Computer Society (Jul 2017).
  \doi{10.1109/CVPR.2017.694}

\bibitem{kendall2015posenet}
Kendall, A., Grimes, M., Cipolla, R.: Pose{N}et: A convolutional network for
  real-time 6-{DOF} camera relocalization. In: Proceedings of the 2015
  International Conference on Computer Vision. pp. 2938--2946. IEEE Computer
  Society (Dec 2015). \doi{10.1109/ICCV.2015.336}

\bibitem{kingma2015adam}
Kingma, D.P., Ba, J.: Adam: {A} method for stochastic optimization. In: Bengio,
  Y., LeCun, Y. (eds.) Proc.~of the International Conference on Learning
  Representations ({ICLR}) (2015)

\bibitem{kneip2011novel}
Kneip, L., Scaramuzza, D., Siegwart, R.: A novel parametrization of the
  perspective-three-point problem for a direct computation of absolute camera
  position and orientation. In: Proceedings of the 2011 Conference on Computer
  Vision and Pattern Recognition. pp. 2969--2976. IEEE, IEEE Computer Society
  (Jun 2011)

\bibitem{lee2019meta}
Lee, K., Maji, S., Ravichandran, A., Soatto, S.: Meta-learning with
  differentiable convex optimization. In: Proc.~of the {IEEE} Conference on
  Computer Vision and Pattern Recognition ({CVPR}). IEEE Computer Society
  (2019)

\bibitem{lepetit2009epnp}
Lepetit, V., Moreno-Noguer, F., Fua, P.: {EPnP}: An accurate {O(n)} solution to
  the {PnP} problem. International Journal of Computer Vision  \textbf{81}(2),
  155--166 (2009)

\bibitem{li2018megadepth}
Li, Z., Snavely, N.: Mega{D}epth: Learning single-view depth prediction from
  internet photos. In: Proc.~of the {IEEE} Conference on Computer Vision and
  Pattern Recognition ({CVPR}). pp. 2041--2050. IEEE Computer Society (2018)

\bibitem{liu2019deep}
Liu, L., Campbell, D., Li, H., Zhou, D., Song, X., Yang, R.: Learning
  {2D}--{3D} correspondences to solve the blind perspective-n-point problem.
  Tech. rep., Australian National University (arXiv:2003.06752) (Mar 2019)

\bibitem{moreno2008pose}
Moreno-Noguer, F., Lepetit, V., Fua, P.: Pose priors for simultaneously solving
  alignment and correspondence. In: Forsyth, D., Torr, P., Zisserman, A. (eds.)
  Proceedings of the 2008 European Conference on Computer Vision. pp. 405--418.
  Springer, Springer (Oct 2008)

\bibitem{qi2017pointnet}
Qi, C.R., Su, H., Mo, K., Guibas, L.J.: {PointNet:} deep learning on point sets
  for {3D} classification and segmentation. In: Proc.~of the {IEEE} Conference
  on Computer Vision and Pattern Recognition ({CVPR}). pp. 652--660. IEEE, IEEE
  Computer Society, Honolulu, USA (Jul 2017)

\bibitem{santacruz2018visual}
{Santa Cruz}, R., Fernando, B., Cherian, A., Gould, S.: Visual permutation
  learning. {IEEE} Trans.~on Pattern Analysis and Machine Intelligence ({PAMI})
   \textbf{41}(12),  3100--3114 (2019)

\bibitem{sattler2017efficient}
Sattler, T., Leibe, B., Kobbelt, L.: Efficient effective prioritized matching
  for large-scale image-based localization. IEEE Transactions on Pattern
  Analysis and Machine Intelligence  \textbf{39}(9),  1744--1756 (Sep 2017).
  \doi{10.1109/TPAMI.2016.2611662}

\bibitem{sattler2019understanding}
Sattler, T., Zhou, Q., Pollefeys, M., Leal-Taixe, L.: Understanding the
  limitations of {CNN}-based absolute camera pose regression. In: Proc.~of the
  {IEEE} Conference on Computer Vision and Pattern Recognition ({CVPR}). IEEE
  Computer Society (Jun 2019)

\bibitem{schonberger2018semantic}
Sch{\"o}nberger, J.L., Pollefeys, M., Geiger, A., Sattler, T.: Semantic visual
  localization. In: Proc.~of the {IEEE} Conference on Computer Vision and
  Pattern Recognition ({CVPR}). pp. 6896--6906. IEEE Computer Society (2018)

\bibitem{schoenberger2016sfm}
Sch\"{o}nberger, J.L., Frahm, J.M.: Structure-from-motion revisited. In:
  Proc.~of the {IEEE} Conference on Computer Vision and Pattern Recognition
  ({CVPR}). IEEE, IEEE Computer Society (2016)

\bibitem{sinkhorn1967diagonal}
Sinkhorn, R.: Diagonal equivalence to matrices with prescribed row and column
  sums. The American Mathematical Monthly  \textbf{74}(4),  402--405 (1967)

\bibitem{stackelberg2011market}
von Stackelberg, H., Bazin, D., Urch, L., Hill, R.R.: Market structure and
  equilibrium. Springer (2011)

\bibitem{svarm2016city}
Sv{\"a}rm, L., Enqvist, O., Kahl, F., Oskarsson, M.: City-scale localization
  for cameras with known vertical direction. IEEE Transactions on Pattern
  Analysis and Machine Intelligence  \textbf{39}(7),  1455--1461 (2016)

\bibitem{walch2017image}
Walch, F., Hazirbas, C., Leal-Taixe, L., Sattler, T., Hilsenbeck, S., Cremers,
  D.: Image-based localization using lstms for structured feature correlation.
  In: Proc.~of the International Conference on Computer Vision ({ICCV}). pp.
  627--637. IEEE Computer Society (2017)

\bibitem{wu20153d}
Wu, Z., Song, S., Khosla, A., Yu, F., Zhang, L., Tang, X., Xiao, J.: {3D}
  {S}hape{N}ets: A deep representation for volumetric shapes. In: Proc.~of the
  {IEEE} Conference on Computer Vision and Pattern Recognition ({CVPR}). pp.
  1912--1920. IEEE Computer Society (2015)

\bibitem{yi2018learning}
Yi, K.M., Trulls, E., Ono, Y., Lepetit, V., Salzmann, M., Fua, P.: Learning to
  find good correspondences. In: Proc.~of the {IEEE} Conference on Computer
  Vision and Pattern Recognition ({CVPR}). pp. 2666--2674. IEEE Computer
  Society (2018)

\end{thebibliography}

\end{document}